\documentclass[conference,a4paper]{IEEEtran}
\usepackage[left=1.62cm,right=1.62cm,top=1.9cm]{geometry}

\IEEEoverridecommandlockouts

\usepackage{graphicx}
\usepackage{subfigure}
\usepackage{siunitx}
\usepackage{xcolor}
\usepackage{algorithm}
\usepackage{comment}
\usepackage{varwidth}
\usepackage{subfig} 
\usepackage{algpseudocode}
\usepackage{enumitem}
\usepackage{caption}
\usepackage{balance}
\usepackage[backend = bibtex,
			style = numeric,
			sorting = none,
			]{biblatex}
\bibliography{references.bib, caleb-zotero.bib}

\usepackage{url}
\usepackage{multirow}

\newcommand{\eg}[1]{}
\renewcommand{\eg}[1]{(e.g. {#1})}

\newcommand{\ie}[1]{}
\renewcommand{\ie}[1]{(i.e. {#1})}

\newcolumntype{L}[1]{>{\raggedright\let\newline\\\arraybackslash\hspace{0pt}}m{#1}}
\newcolumntype{C}[1]{>{\centering\let\newline\\\arraybackslash\hspace{0pt}}m{#1}}

\begin{document}

\title{Survey on Computer Vision Techniques for Internet-of-Things Devices}

\author{\IEEEauthorblockN{Ishmeet Kaur\thanks{Both authors, Ishmeet Kaur and Adwaita Janardhan Jadhav, are currently at Apple Inc.} and Adwaita Janardhan Jadhav}
\IEEEauthorblockA{
}}

\maketitle

\begin{abstract}

Deep neural networks (DNNs) are state-of-the-art techniques for solving most computer vision problems.
DNNs require billions of parameters and operations to achieve state-of-the-art results. This requirement makes DNNs extremely compute, memory, and energy-hungry, and consequently difficult to deploy on small battery-powered Internet-of-Things (IoT) devices with limited computing resources. Deployment of DNNs on Internet-of-Things devices, such as traffic cameras, can improve public safety by enabling applications such as automatic accident detection and emergency response.
Through this paper, we survey the recent advances in low-power and energy-efficient DNN implementations that improve the deployability of DNNs without significantly sacrificing accuracy. In general, these techniques either reduce the memory requirements, the number of arithmetic operations, or both. 
The techniques can be divided into three major categories: (1) neural network compression, (2) network architecture search and design, and (3) compiler and graph optimizations. 
In this paper, we survey both low-power techniques for both convolutional and transformer DNNs, and summarize the advantages, disadvantages, and open research problems.


\end{abstract}


\section{Introduction}

Deep Neural Networks (DNNs) are important tools in computing. They are heavily used in fields like computer vision. 
Object detection, classification, and segmentation~\cite{ZZ, xliu} applications are often built with DNNs. 
DNNs are not just simple neural networks with a few hundred neurons, they 
are made of hundreds of layers with billions of trainable parameters. Having a large number of trainable parameters is the reason why DNNs are computationally expensive. The latest Vision Transformer (ViT) needs over 100 billion operations to perform image classification on a single image~\cite{han2016}. Similarly, the recently published YOLOv8 architecture requires over 68 million parameters to perform accurate object detection. Storing millions of parameters and/or performing billions of operations makes DNNs difficult to deploy on Internet-of-Things (IoT) devices that are usually characterized by low-cost hardware and limited battery lives.



\begin{figure}[t!]
\centering
\subfigure[]{\includegraphics[width = 0.49\textwidth]{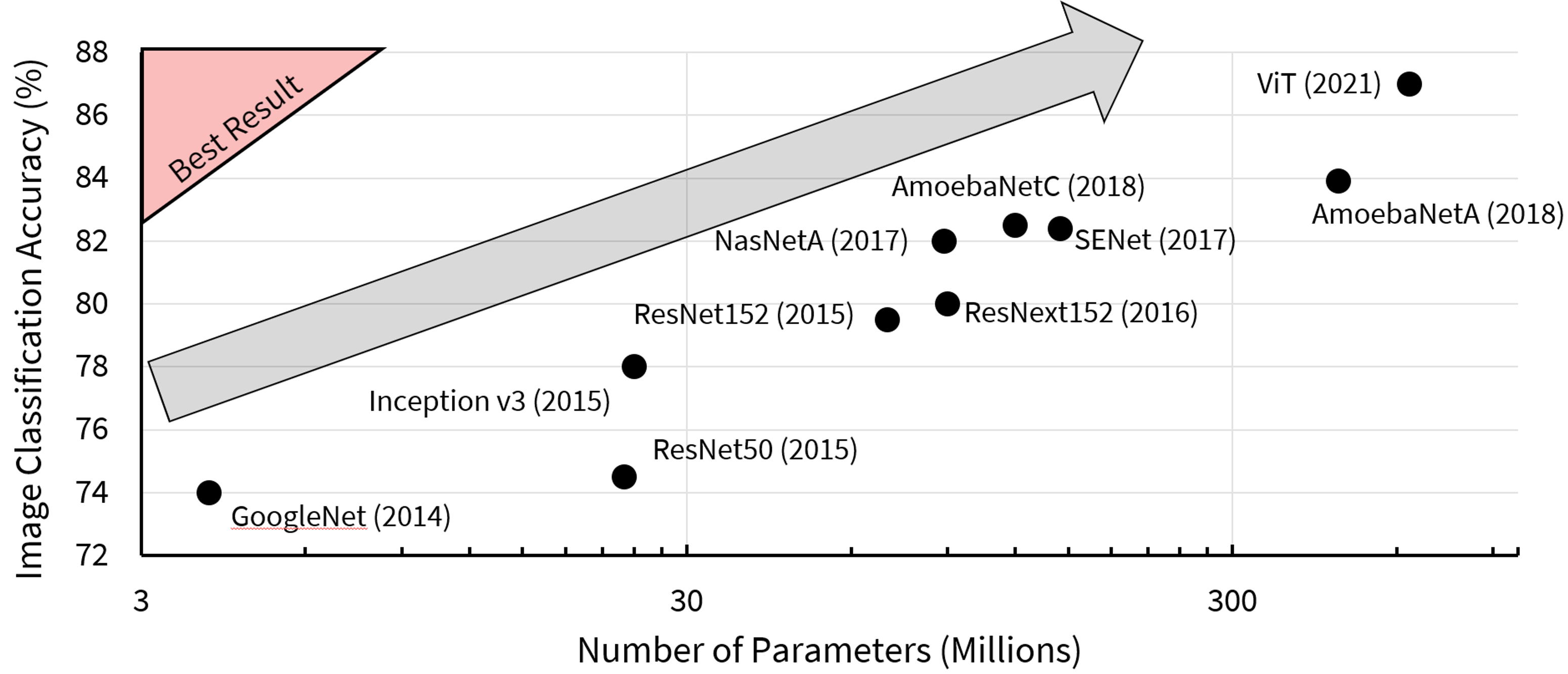}
        }
\subfigure[]{\includegraphics[width = 0.49\textwidth]{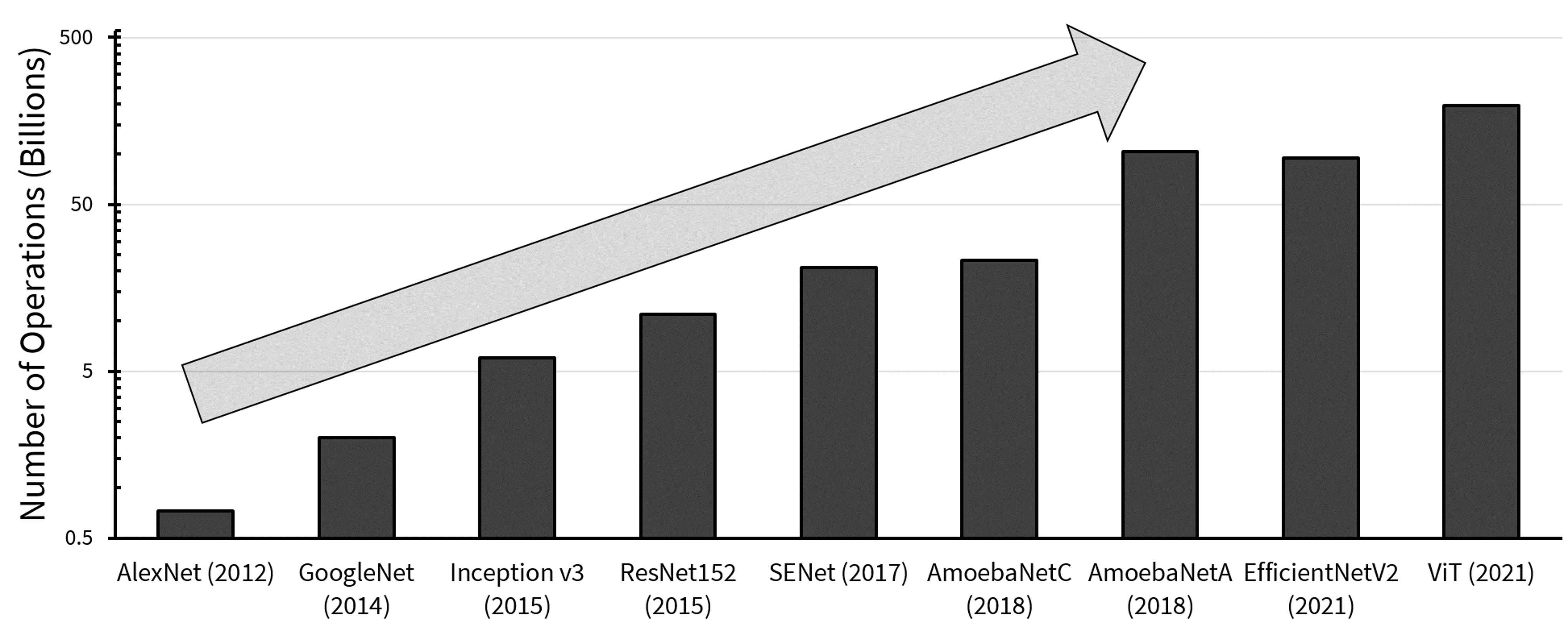}}

\caption{Over time DNNs have become larger and more complex. (a)~Comparison of the number of parameters and test accuracy on the ImageNet-2012 image classification task. In most cases, to achieve higher accuracy, more parameters are required. (b)~Chart comparing the number of operations required to classify one image from the ImageNet-2012 dataset.  Please note the logarithmic scale on the axes.}
\label{fig:one}
\end{figure}

The general trend in computer vision research is to make DNNs larger and more complex in order to improve accuracy. Fig.~\ref{fig:one}(a) and Fig.~\ref{fig:one}(b) show trends in the number of arithmetic operations and the number of parameters required in state-of-the-art DNN models over years, respectively. In Fig.~\ref{fig:one}(a), we can see that from GoogleNet (2014) to ViT (2021), the number of parameters has increased by a factor of 100. Fig.~\ref{fig:one}(b) shows that in a similar time range, the number of arithmetic operations has also increased from $\sim$750 million per image to $\sim$200 billion per image. This continued trend of increasing the DNN complexity makes DNNs harder to deploy on resource-constrained IoT devices.

Deploying DNNs on IoT devices will open up many new computer vision applications. Large-scale computer vision can be performed on cameras in areas without reliable network coverage, such as forests and mountains, to perform tasks like automatic wildlife detection and tracking. Similarly, drones and unmanned aerial vehicles can automatically navigate themselves to aid in disaster relief operations. To enable such applications, researchers have developed many techniques to identify and remove redundancies from DNNs to increase their efficiency without sacrificing accuracy. Since most IoT devices are only equipped with CPUs and limited memory, more research is required to optimize DNNs and reduce their energy footprint further.

\begin{table*}[t!]
\centering
\begin{tabular}{|C{2.7cm}|L{4.7cm}|L{3.7cm}|L{4.5cm}|}
\hline
Approach & \multicolumn{1}{c|}{Summary} & \multicolumn{1}{c|}{Advantages} & \multicolumn{1}{c|}{Open Research Areas} \\ \hline
Neural Network Compression & 
Reduces the number of bits to represent the weights, biases, and activations. & Low accuracy losses in most cases. Can be applied to different types of models. & Most general purpose hardware does not support custom bit formats or sparse matrix multiplications. \\ \hline
Architecture Search and Design & New layer types and DNN structures that are efficient for specific tasks. & State-of-the-art accuracy. & Requires an expert to design DNNs or requires high training costs. \\ \hline
Compiler and Graph Optimizations & Automatically detect and optimize common DNN operations. & Leads to better utilization of hardware. & Not applicable to all operations. \\ \hline
\end{tabular}
\caption{Comparison of different techniques for performing low-power computer vision.}
\label{tab:comp}
\end{table*}



This paper surveys the influential research on energy-efficient DNNs specifically designed for IoT devices.
We use the results reported in the existing literature to analyze and compare the different techniques.
To the best of our knowledge, this is the first paper to discuss how the following low-power techniques can be applied to optimize inference for both convolutional and transformer-based DNNs.  TABLE~\ref{tab:comp} summarizes the findings of this survey. 

\begin{enumerate}[noitemsep,nolistsep]
    \item Neural Network Compression: These techniques reduce the workload size by reducing the number of DNN parameters or the number of bits allocated to the DNN parameters/activations. Some of these techniques also use matrix decomposition techniques to further decrease the memory requirement and the number of operations.
    
    \item Architecture Search and Design: These techniques use new types of layers, interconnections between layers, and filter designs to build DNN architectures that are more optimized for certain types of tasks.
    
    \item Compiler and Graph Optimizations: These techniques optimize the tensor memory layouts, propose new algorithms for performing matrix multiplications, or fuse operations to reduce the number of memory accesses.
    
\end{enumerate}

\section{Neural Network Compression}
\label{sec:compress}

\subsection{Quantization}

By default, most deep learning frameworks such as PyTorch and Jax represent floating-point values using 32 bits. These 32-bit floating point values may be needed during training to maintain high precision. However, during inference, a technique called quantization can be used to reduce the number of bits, and consequently reduce the number of memory operations required by a DNN. Quantization techniques are methods to map 32-bit floating point values to lower precision formats such as 16-bit floating point or 8-bit integer~\cite{wu_quantized_2016}. 

Several hardware processors, including the NVIDIA Jetson Nano and Raspberry Pi 4B support low-precision arithmetic. On these processors, 8-bit integer operations can be close to 10$\times$ faster than 32-bit floating point operations. Furthermore, research has shown that integer operations can consume close to 14$\times$ less energy than floating point operations. 

\begin{figure}[t!]
\centering
\includegraphics[width = 0.49\textwidth]{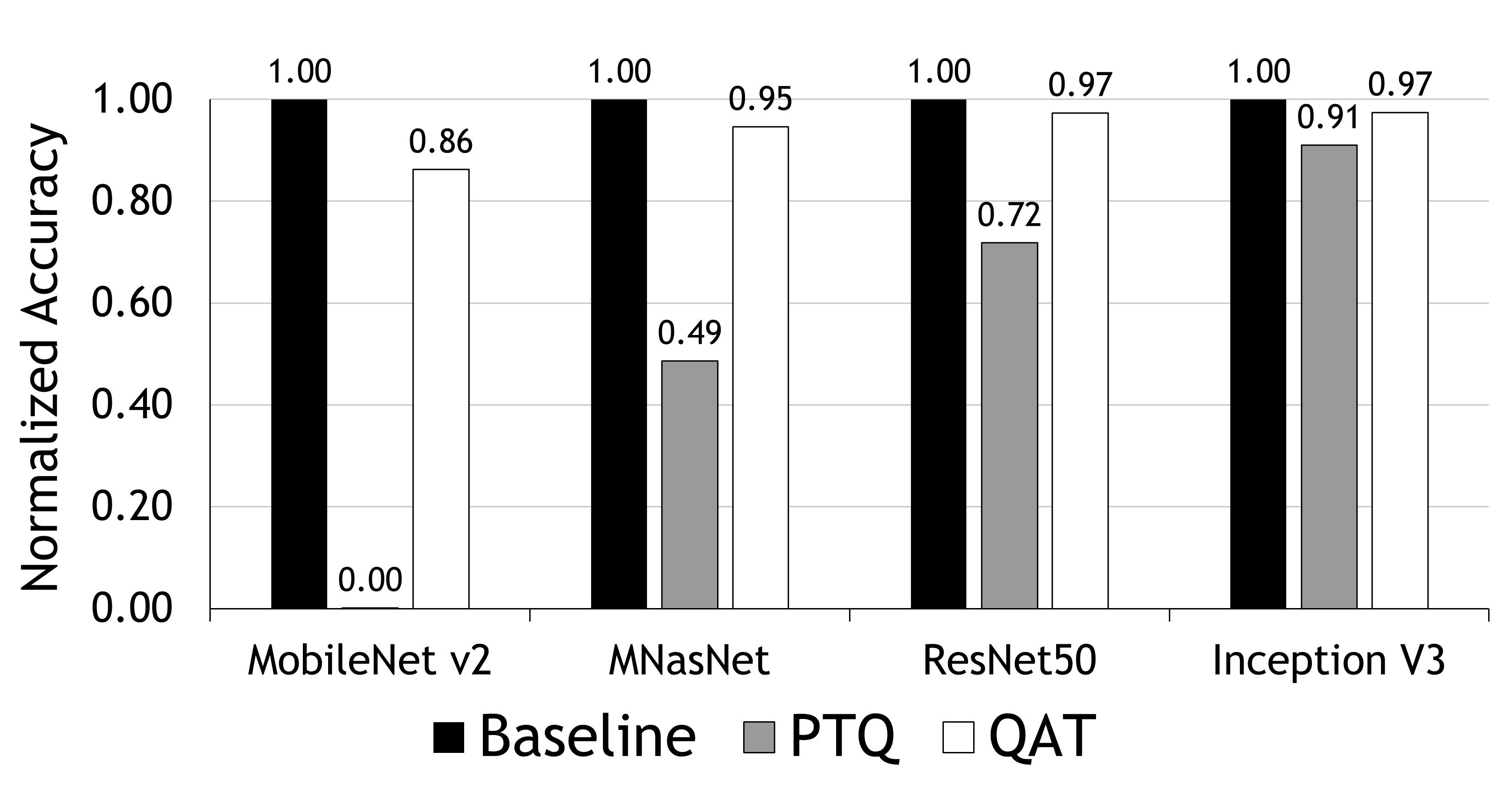}
\caption{Comparison of convolutional DNN accuracy when performing 4-bit quantization. The accuracy is normalized to the baseline accuracy. Quantization Aware Training results in smaller accuracy losses than Post Training Quantization.}
\label{fig:two}
\end{figure}

Quantization Aware Training (QAT) and Post-Training Quantization (PTQ) are the two main methods to perform quantization. QAT methods perform quantization during DNN training to reduce accuracy losses. Generally, these techniques are more accurate but have expensive training processes~\cite{han_deep_2015}. As the name suggests, PTQ methods perform quantization once training is complete. These techniques use little or no labeled data to perform quantization. Thus, these techniques do not increase the training complexity but are generally less accurate.
The difference in accuracy between QAT and PTQ is seen in Fig~\ref{fig:two}. We see that for most convolutional DNNs, QAT outperforms PTQ consistently~\cite{krishnamoorthi_quantizing_2018}. The accuracy vs. complexity tradeoff of both QAT and PTQ techniques can be tuned by adjusting the symmetry, granularity, and uniformity. This creates a complex design landscape and often requires experts to carefully tune the different hyper-parameters.

A significant amount of research has focused on using quantization to improve the efficiency of transformer-based DNN architectures. The work by Kim et al.~\cite{DBLP:journals/corr/abs-2101-01321}. Special quantization functions, such as power-of-two and log-int-softmax, are used to approximate the GELU activation function, Softmax, and Layer Norm into integer arithmetic~\cite{DBLP:journals/corr/abs-2111-13824}. 
Liu et al.~\cite{liu2021post} 
present a method to use PTQ to reduce the parameter size of ViT from 1,228 GB to 231 GB with only 1\% accuracy loss on the ImageNet dataset. This result indicates that Transformer architectures are less sensitive to PTQ when compared with convolutional DNNs.

\begin{figure}[b!]
\centering
\includegraphics[width=0.49\textwidth]{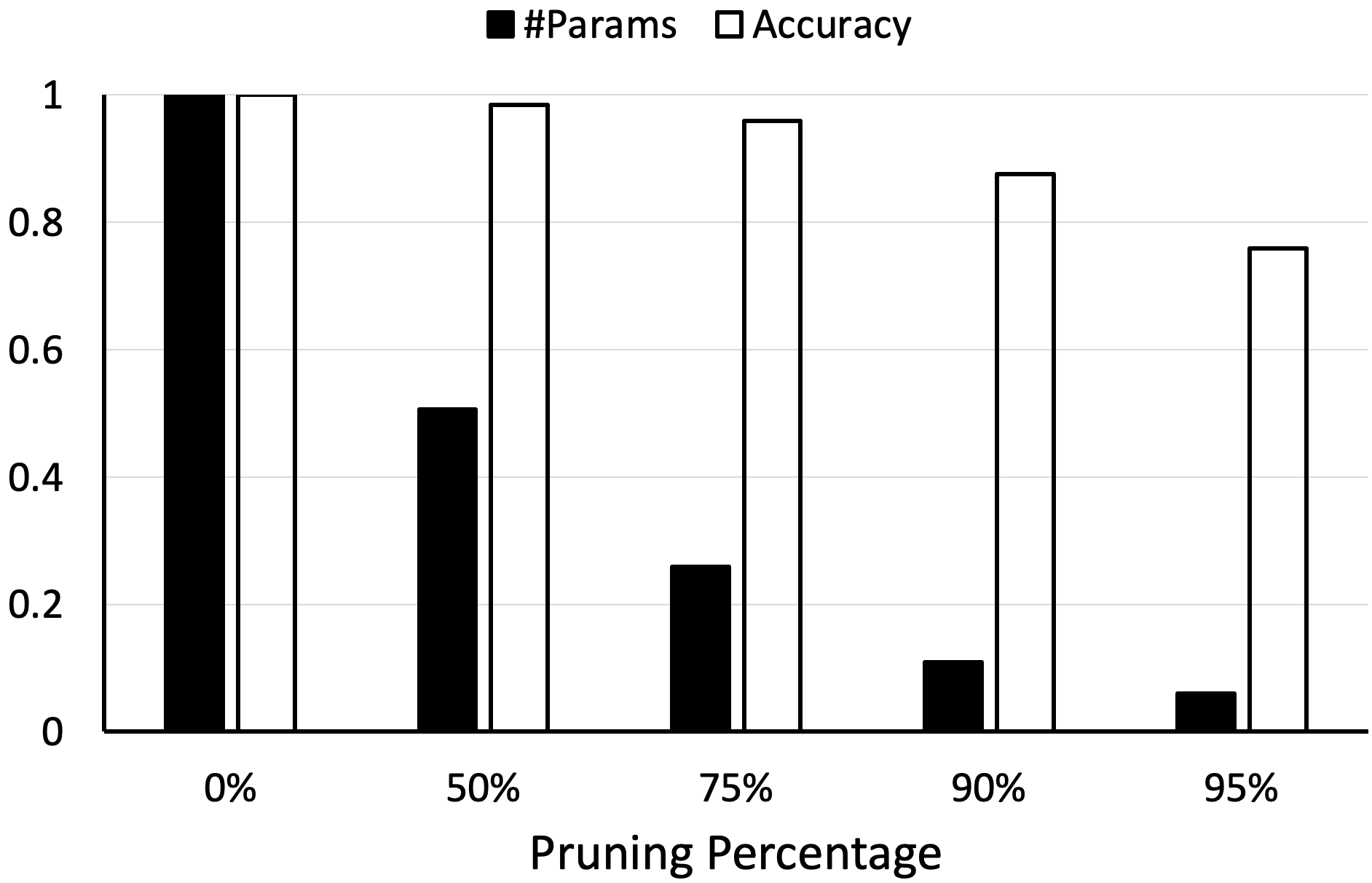}
\caption[List of figure caption goes here]{Comparison of MobileNet V2 normalized accuracy and normalized number of parameters with different levels.}
\label{fig:pruning}
\end{figure}

\subsection{Pruning}

The latest DNNs have billions of weights and biases. However, research has shown that DNNs are often over-parameterized, and not all the weights are required to achieve high accuracy. DNN Pruning is a technique that can reduce the size of DNNs by removing some of these redundant weights and biases.

Pruning was first proposed in works called ``Optimal Brain Damage''~\cite{lecun_optimal_1990} and  ``Optimal Brain Surgeon'' published in the 1990s. These works suggested that DNNs can be simplified by removing the ``least important'' weights, without sacrificing the overall accuracy. Over the years, many researchers have focused their efforts on designing newer and more accurate methods to quantify the importance of the DNN weights. The pruning techniques can be classified into three categories: (1) value-based, (2) accuracy-based, and (3) iterative.

Value-based methods prune DNNs by rounding all the small DNNs weights to zero. However, subsequent research has shown that this pruning technique is suitable for a small range of compression ratios~\cite{han_deep_2015}. As an improvement on value-based pruning techniques, researchers have proposed accuracy-based pruning techniques. In these techniques, the importance of the weights is estimated by measuring the DNN accuracy losses when the weights are removed. If the accuracy losses are significant, the weight is considered to be important. Once all the weights have been assigned an importance score, the bottom-ranked weights can be removed. The limiting factor in these techniques is the computational costs associated with quantifying the importance of every weight. For DNNs with billions of parameters, these computational costs are prohibitively high. As a compromise, future researchers could look into the possibility of creating a few groups of similar weights and quantify the importance of the weight groups.

Iterative pruning is a different pruning paradigm that includes the pruning process in the training algorithm. Iterative pruning was first proposed by Han et al.~\cite{han_deep_2015}. In these techniques, a regularization function is computed along with the loss during training. The regularization function forces the backpropagation algorithm to assign a non-zero value to a weight if the weight is activated multiple times and to assign a zero to the less frequently used weights. Because the weights are pruned iteratively in each training epoch, the DNN can retain accuracy to a greater extent.

\begin{figure}[b!]
\centering
\includegraphics[width=0.49\textwidth]{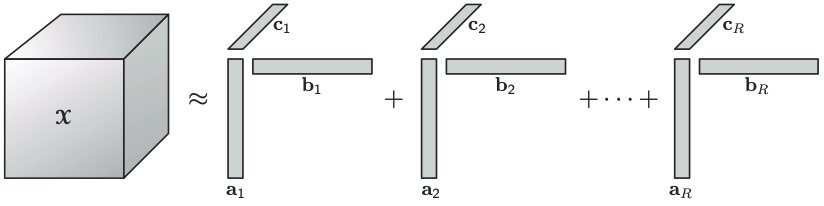}
\caption[List of figure caption goes here]{CP decomposition of a convolution kernel in a DNN. This decomposition factorizes a tensor into a sum of component rank-one tensors. Tensor decomposition reduces the memory requirement and number of operations significantly. Figure source: Kolda et al.~\cite{kolda_tensor_2009}}
\label{fig:factor}
\end{figure}

\subsection{Layer Decomposition}

The most accurate DNNs have large weight matrices~\cite{simonyan_very_2015}. For example, the popular VGG-16 DNN has six layers with shape $3\times 3\times 512 \times 512$. 
Results in Kolda et al.~\cite{kolda_tensor_2009} show that large matrices can be decomposed into smaller low-rank matrices. An example of the decomposition is depicted in Fig.~\ref{fig:factor}.
These low-rank matrices reduce the peak memory requirements of the DNN, and thus make it easier to fit within the limited resource constraints of IoT devices~\cite{jaderberg_speeding_2014, denton_exploiting_2014}.

There are many different decomposition techniques that can be used. Some examples include Singular Value Decomposition (SVD), Canonical Polyadic Decomposition (CPD), and Tucker-2 Decomposition. Many researchers have noted that there are non-negligible accuracy losses when using existing decomposition techniques on pre-trained DNNs~\cite{tai_convolutional_2016, lebedev_speeding-up_2015}. Moreover, it has also been shown that SVD often leads to lower computer vision accuracy than Tucker-2 or CPD~\cite{lebedev_speeding-up_2015}. Understanding why some decomposition techniques perform better than others is an open problem and requires research.

Similar to iterative pruning, a regularization function can be used to train the DNN weights to meet low-rank requirements. Including the decomposition process in the training algorithm ensures that the accuracy losses are minimal. Alvarez et al.~\cite{alvarez_compression-aware_2017} and Xiong et al.~\cite{xiong_trp_2020} are examples of such techniques. These techniques have been shown to have efficiency gains, but require significant resources during training time.

Applications of decomposition techniques have also been explored for transformer architectures. Metha et al.~\cite{mehta2020low} 
show that the multi-head attention unit can be decomposed to reduce its complexity. These results are particularly important as the attention sequence lengths become longer for video processing applications. Lan et al.~\cite{lanalbert}
showcase similar performance improvements with decomposition on the BERT model.

\section{Architecture Search and Design}
\label{sec:design}

\begin{figure}[t!]
\centering
\includegraphics[width=0.49\textwidth]{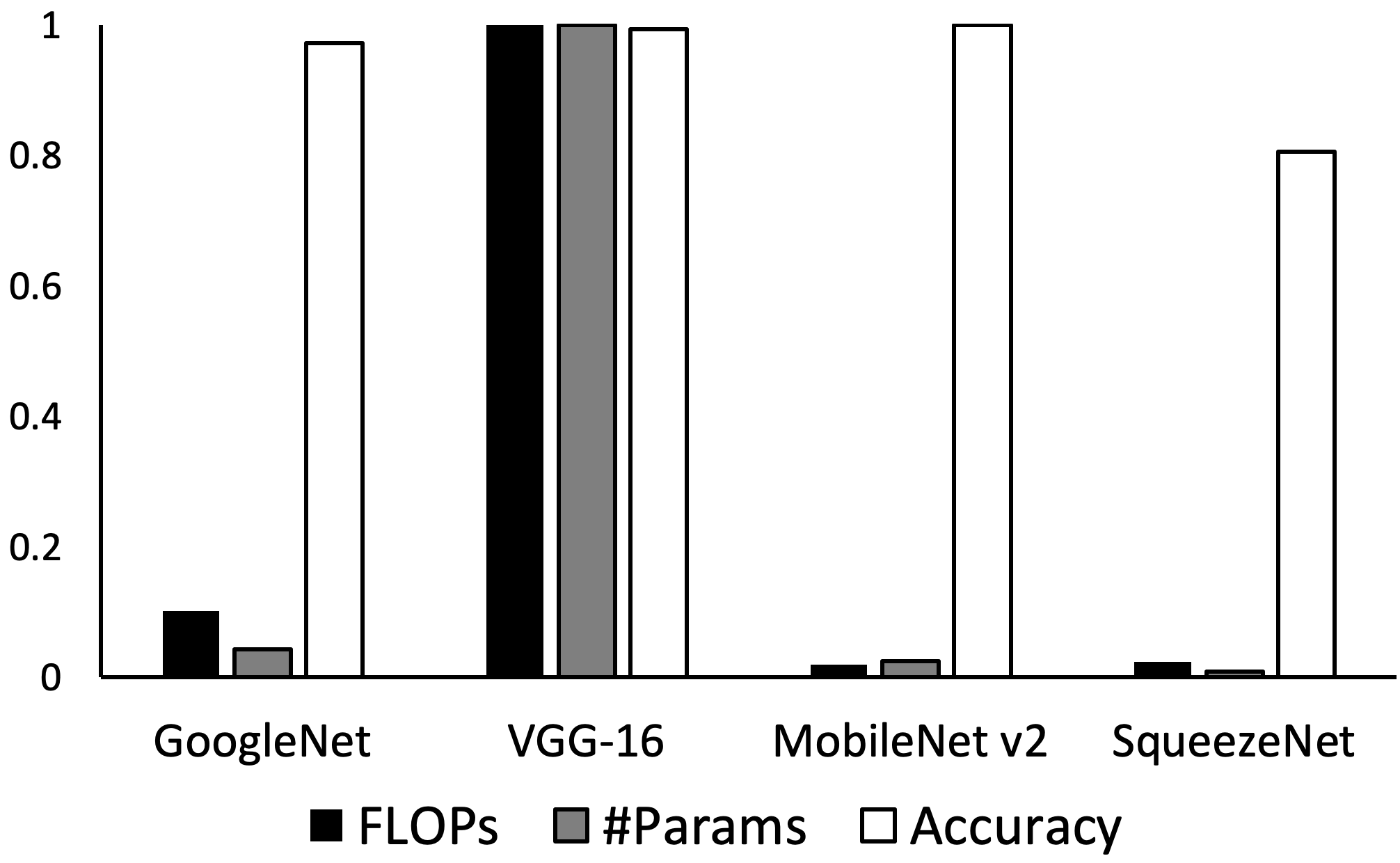}
\caption{Comparison of the normalized FLOPs, \#parameters, and accuracy between four representative DNN architectures. MobileNet and SqueezeNet maintain classification accuracy but only require a fraction of the computational resources.}
\vspace{-0.1in}
\label{fig:design}
\end{figure}

The standard convolution operation is implemented as a tensor multiplication between the parameter tensor and the input tensor.
The same convolution kernel is strided over the entire image to generate the output. Therefore, the number of arithmetic operations performed during convolution is quadratic with the size of the convolution kernel. This knowledge is the motivating reason for many novel DNN architecture designs that use small convolutional kernels. 
The two most famous architectures that use small convolutional kernels are MobileNet~\cite{sandler_mobilenetv2_2018} and SqueezeNet~\cite{iandola_squeezenet_2016}. These DNNs use a new operation called the depthwise separable convolutions to beat the state-of-the-art low-power computer vision solutions. The depthwise separable convolution breaks up a single convolutional kernel into two sub-tasks: (1)~depthwise convolution: applies a single convolutional channel filter on each channel of the input, and (2)~pointwise convolution: performs $1 \times 1$ convolutions to combine the depthwise convolutions. Fig.~\ref{fig:design} shows the benefit of using MobileNet and SqueezeNet over existing DNN architectures. These techniques are can also be used with the previously described pruning and quantization techniques to further increase the efficiency. 
 
EfficientNet and EfficientDet~\cite{tan_efficientdet_2020} provide guidebooks for building different DNN architectures. They showcase the optimal methods to perform layer scaling, width scaling, and resolution scaling. These guides enable the design of compact and accurate DNNs for different resource budgets.

Because there are many different DNN architectures, it can be difficult to select the best DNN for a given task. 
Network Architecture Search (NAS) is a technique that solves this problem by automating architecture selection. NAS uses a ``controller'' to propose DNN architectures. Reinforcement learning is used to evaluate these proposed DNN architectures and then propose another DNN architecture. AmoebaNet~\cite{amoeba} showcases the ability of NAS to obtain state-of-the-art results.

Tan et al.~\cite{MNas} propose MNasNet as a new NAS technique that can build efficient and IoT-friendly DNN models. MNasNet uses two objectives in the reinforcement learning reward function: (1) accuracy and (2) latency.
However, MNasNet requires over 4.6 GPU years to train one DNN architecture for image classification on the ImageNet dataset. This amount of computing resource requirements during training is a challenge that needs to be overcome.

Cai et al.~\cite{proxyless} propose ProxylessNAS, a new NAS technique with the goal of reducing the training overheads.  Proxyless-NAS uses an estimator function to measure the latency and a novel path-level pruning algorithm to reduce the number of candidate architectures. ProxylessNAS reduces the training time of efficient DNNs from 4.6 GPU years to approximately 300 GPU hours. FBNet~\cite{FB} is another technique that reduces the training overheads. FBNet uses a proxy task, i.e., optimizing over a smaller dataset to evaluate candidate architectures. These architecture search techniques are compatible with both convolutional and transformer-based architectures~\cite{liu2022neural}. 

Future work on architecture search should focus on algorithms that can find architecture designs that contain different building blocks from the existing human-made DNNs. The existing solutions only try different combinations of well-known ``architectural blocks''. Breaking this dependency may allow us to build new architectures.

\begin{figure*}[t!]
\centering
\includegraphics[width=0.99\textwidth]{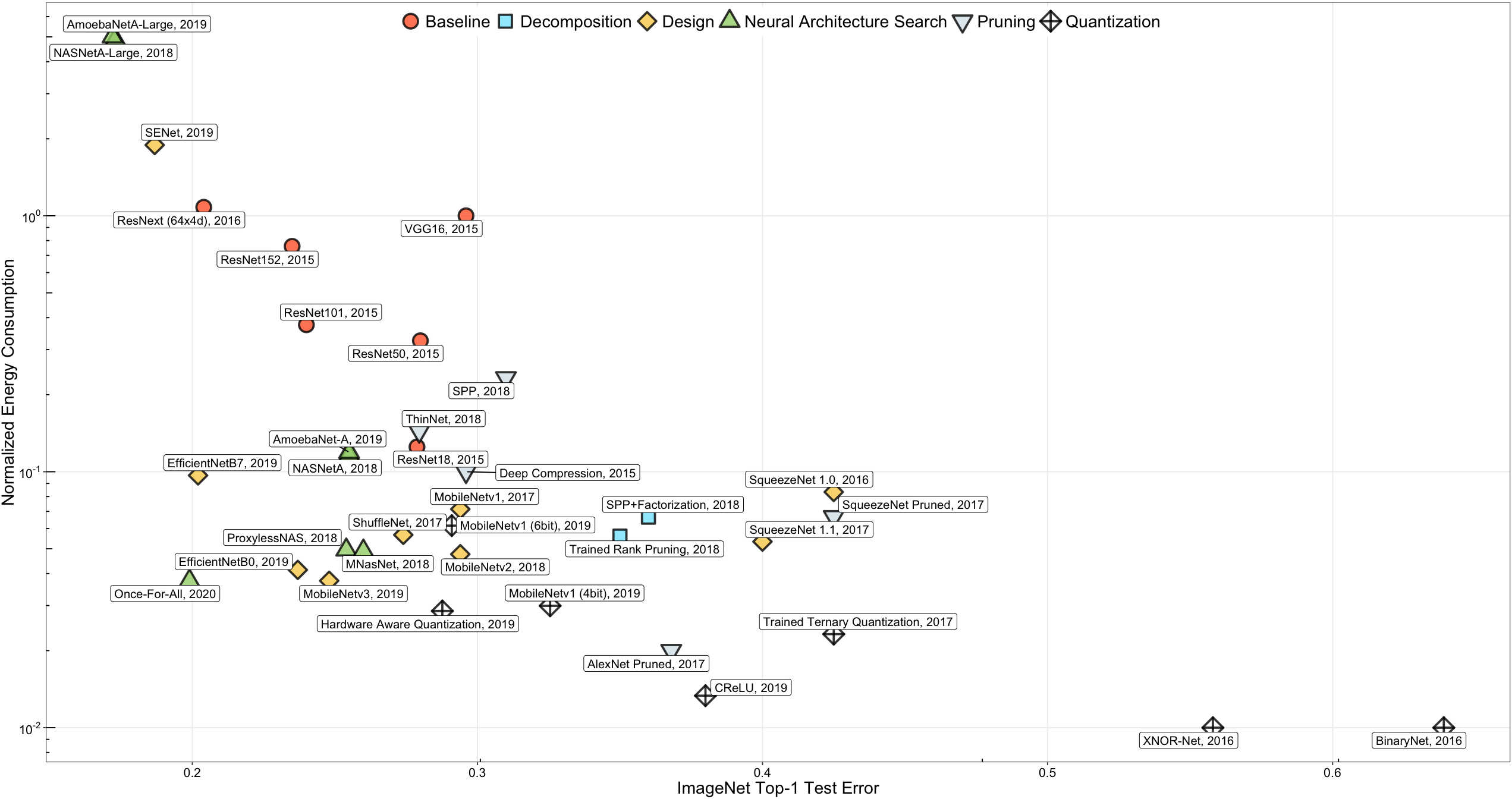}
\caption{Comparison of energy consumption and accuracy of different computer vision optimizations. The data-points closer to the bottom right are considered to be the best, i.e., low energy but high accuracy. Circles denote the baseline DNNs deployed without any optimizations. Different techniques can be combined with quantization and pruning to vary the accuracy-efficiency tradeoff. Best viewed in color on a screen.}
\label{fig:survey}
\vspace{-0.1in}
\end{figure*}
\begin{figure}[t!]
\centering
\includegraphics[width=0.3\textwidth]{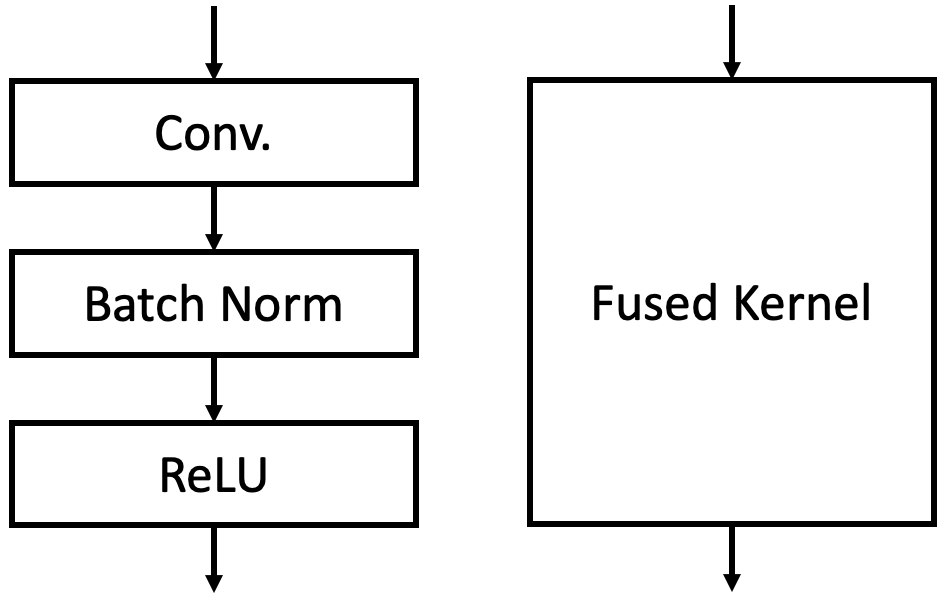}
\caption{Right: unfused convolutional operation that is commonly used in DNNs. Left: Fused kernel that performs the same operation with a single instruction.}
\label{fig:fusion}
\vspace{-0.1in}
\end{figure}

\section{Compiler and Graph Optimizations}

While executing a DNN on any deep learning framework, often similar computations are performed routinely. Researchers have proposed deep learning compilers like TensorRT and XLA that can identify common patterns and optimize them. These complier optimizations include techniques like layer and tensor fusions~\cite{snider2023operator}, kernel tuning~\cite{10.1145/3508391}, and focused convolutions~\cite{tung2022irrelevant}.

In layer fusion techniques, different operations are performed with a single instruction. Layer fusion is depicted in Fig.~\ref{fig:fusion} where the convolution, batch norm, and ReLU operations are fused into a single kernel~\cite{snider2023operator}. Fused kernels reduce the overhead associated with reading and writing the intermediate tensor data for each constituent operations. Kernel auto-tuning techniques~\cite{10.1145/3508391} perform automatic hyper-parameter search to identify the most optimal configurations to meet the required accuracy-efficiency tradeoff. These auto-tuned kernels can vary the tensor precision, network sparsity, batch size, and resolution to vary the DNN's performance. Finally, the focused convolutions modify the General Matrix Multiply (GeMM) algorithm to reduce redundant operations~\cite{tung2022irrelevant}. They focused convolution removes rows from the GeMM input matrix that correspond to ``irrelevant'' pixels to avoid performing convolutions on those inputs. Such techniques are particularly useful in situations where the image background (e.g., trees, sky, buildings) are not relevant to the computer vision task.

Most existing compilers are not general enough to fuse or optimize any DNN architecture. They are well suited for some architectures such as MobileNet~\cite{sandler_mobilenetv2_2018} or BERT~\cite{lanalbert}. Future work on deep learning compilers should focus on generality to improve their use. Furthermore, although compilation is an one-time expense, the resources required for compilation can be significant. More work on optimizing the compilation time will help increase their deployability on IoT devices.

\section{Discussion and Conclusion}

\subsection{Summary of Findings}

This section summarizes our findings related to low-power DNN optimizations. Fig.~\ref{fig:survey} plots the accuracy and energy consumption of different techniques when deployed on IoT devices. The baseline techniques without any optimizations are marked with red circles. Efficient architecture designs such as MobileNetV3~\cite{howard_searching_2019} and EfficientNet~\cite{tan_efficientdet_2020} can lower the energy consumption without sacrificing accuracy. Aggressive 1-bit quantization techniques have been proposed in XNORNet~\cite{xnor} and BinaryNet~\cite{BinNet}, however, these techniques suffer from significant accuracy losses. More research is required to understand how these 1-bit techniques can be used without accuracy losses. More finegrained control over the accuracy vs efficiency tradeoff may be possible by combining different types of techniques. However, most combinations have not yet been explored exhaustively and is an open area of research.




\subsection{Proposed Evaluation Metrics}

Based on our findings, we propose new evaluation metrics for future researchers to test the deployability of their solutions on low-power IoT devices.
Most DNN-based computer vision techniques are evaluated on just accuracy~\cite{he_deep_2016}. In this section, we present some additional metrics that should be used to evaluate future low-power DNN techniques.

\begin{itemize}
\item The memory requirement to process a single image should be reported. Prior research has shown that memory operations require up to $200\times$ more energy than arithmetic operations~\cite{nazemi_nullanet_2018, lpirc}. Techniques that require less memory, or that are more cache-efficient have advantages on IoT devices. The number of parameters is usually a good proxy for the number of memory operations. However, there are some exceptions to this rule. For example, techniques like MobileNet v1 have a small number of parameters but have low arithmetic intensity (i.e., perform few arithmetic operations for every memory operation). In such cases, the technique may perform many memory operations even with a small number of parameters. As another example, techniques deploying quantization may have more parameters, but require fewer bits per parameter. In such cases, researchers should include results on the scaling of energy consumption with different model sizes and quantization formats.
\item The number of operations is another important metric to compare the efficiency of different techniques. Although each arithmetic operation requires little energy~\cite{nazemi_nullanet_2018}, when DNNs perform billions of operations the energy costs add up. Thus fewer operations are more useful when deploying on resource constrained hardware. Research has found that fully-connected layers require fewer operations than convolutional layers~\cite{simonyan_very_2015}. However, fully-connected layers have more parameters and may require more memory. When designing low-power DNNs, it is important to consider the advantages and disadvantages of the layers based on the available hardware. In most cases, it is important to measure the energy consumption to understand the impact of design decisions.
\end{itemize}

\subsection{Conclusion}

DNNs have gained popularity over the last decade. More recently, their use has become more ubiquitous, and have started to get deployed on low-power IoT devices. However, since most DNNs are designed for high accuracy on server-grade hardware, it is difficult for practitioners to deploy them on battery-powered devices. In this survey, we highlight and summarize the key research topics that help reduce the energy consumption of DNNs so that they can be deployed on IoT devices. We identify that the existing research can be divided into three main categories: (1) neural network compression, (2) network architecture search and design, and (3) compiler and graph optimizations. These techniques have their advantages and disadvantages. We highlight these features and provide recommendations on how these techniques can be improved with continued research.
\printbibliography

\end{document}